\documentclass[10pt, a4paper]{article}
\usepackage{overpic}
\usepackage{multirow}
\usepackage{lrec-coling2024} 
\title{Medical Vision-Language Pre-Training for Brain Abnormalities}

\name{Masoud Monajatipoor, Zi-Yi Dou, Aichi Chien, Nanyun Peng, Kai-Wei Chang} 

\address{UCLA\\
         \{monajati, zdou, aichi\}@ucla.edu\\
         \{violetpeng, kwchang\}@cs.ucla.edu\\}

\abstract{
Vision-language models have become increasingly powerful for tasks that require an understanding of both visual and linguistic elements, bridging the gap between these modalities. In the context of multimodal clinical AI, there is a growing need for models that possess domain-specific knowledge, as existing models often lack the expertise required for medical applications. In this paper, we take \textit{brain abnormalities} as an example to demonstrate how to automatically collect medical image-text aligned data for pretraining from public resources such as PubMed.
In particular, we present a pipeline that streamlines the pre-training process by initially collecting a large brain image-text dataset from case reports and published journals and subsequently constructing a high-performance vision-language model tailored to specific medical tasks. We also investigate the unique challenge of mapping subfigures to subcaptions in the medical domain. We evaluated the resulting model with quantitative and qualitative intrinsic evaluations. The resulting dataset and our code can be found here {\url{https://github.com/masoud-monajati/MedVL_pretraining_pipeline}}
 \\ \newline  
 }

\begin{document}

\maketitleabstract

\section{Introduction}

Readily available open-access datasets in the broad domain~\cite{lin2014microsoft, sharma2018conceptual} have enabled the development of vision-language (VL) models. Typically, researchers pre-train VL models on large image-text data and then fine-tune them for specific downstream tasks. Such a pre-training/fine-tuning paradigm is highly effective for tasks with limited data and therefore is a standard approach for various downstream applications.

On the other hand, the medical domain presents unique challenges due to the scarcity and complexity of its data \cite{brigato2021close}. 
Pre-trained models, trained on general domain datasets, often exhibit reduced effectiveness when applied to a medical domain with limited data, as observed in the cases of chest x-ray analysis \cite{wu2023medklip, monajatipoor2022berthop} and Alzheimer's disease detection \cite{chen2023medblip}. Furthermore, domain-specific data suitable for pre-training is notably limited. Although several domain-specific medical Vision-Language (VL) datasets do exist, such as MIMIC \cite{johnson2019mimic}, CheXNet \cite{rajpurkar2017chexnet}, and datasets for Alzheimer's disease \cite{petersen2010alzheimer}, their creation requires substantial manual curation, involving significant labor and time. Besides, they are presented for specific domains and their knowledge is not transferable to other medical domains. 

In this paper, we take the brain disease domain as an example and propose an automatic pipeline for extracting image-text pairs for pre-training a VL model for specific medical domains.
The pipeline first collects raw image-text pairs from medical sources like PubMed
and prepares aligned image-text pairs which can be leveraged for VL pre-training. 
There are two main challenges with the data we collected. First, medical data in both image and text form inherently possess a complex nature \cite{han2018deep}. 
Moreover, image-caption pairs sourced from PubMed literature and journals often incorporate subfigures and subscaptions, introducing additional intricacies that can pose formidable challenges for the pre-training of VL models. Therefore, there is a unique challenge in medical VL pre-training where we care about the fine-grained alignment between subfigures/subcaptions. Our pipeline
is equipped with subfigure/subcaption aligning designed to enhance multimodal learning. The letters in subfigures and subcaptions are often referred to as "subfigure labels" or "subcaption labels." are used to identify the alignment between subfigures and subcaptions.

\begin{figure*}
\begin{center}
\includegraphics[scale=0.387]{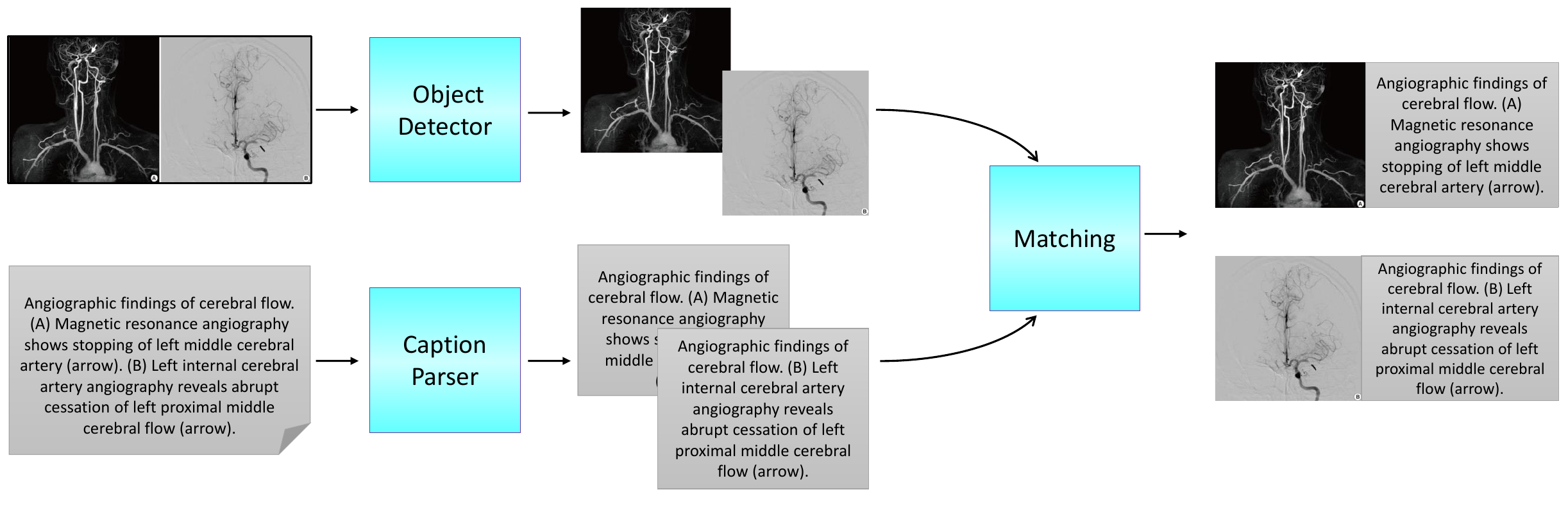} 
\caption{Our main pipeline for matching subfigures/subcaptions. The object detector outputs subfigures while the caption parser parses the caption into subcaptions simultaneously. Then, the module called 'matching' provides us the subfigure/subcaption pairs for the pre-training}
\label{main}
\end{center}
\end{figure*}

Getting the aligned image-text pairs from our pipeline, we pre-train a VL model, BLIP \cite{li2022blip}, on both the original collected data and processed data to analyze the effectiveness of our pipeline. We consider quantitative and qualitative intrinsic evaluations including image-text retrieval and attention visualization. We have observed that the model pre-trained on the processed data has a better multimodal understanding. We believe that our pipeline can be used for other domain-specific medical applications such as Prostate Cancer Diagnosis or Alzheimer's Disease Prediction.

Our work has three main contributions. We identified the challenge in medical VL pre-training by taking brain disease as an example. We built a pipeline that collects domain-specific medical image-text pairs followed by a module for matching subfigures/subcaptions. We release the dataset which is suitable for pre-training a VL model for brain diseases upon the acceptance of our work. Lastly, we pre-trained a VL model (BLIP) \cite{li2022blip} on our dataset, and with quantitative and qualitative analysis, we demonstrate that our pre-trained dataset and model are useful for building VL models in the medical domain.

\begin{table*}
\centering
\begin{tabular}{llllllllll}
\hline
& training schema & \multicolumn{4}{c}{val} & \multicolumn{4}{c}{test}\\
 & & i2t@1 & i2t@10 & t2i@1 & t2i@10 & i2t@1 & i2t@10 & t2i@1 & t2i@10\\
\hline
\multirow{2}{*}{Only PT} & raw data &18.04 & 48.36 & 21.51 & 51.98 & 17.37 & 47.65 & 19.58 & 39.86\\
                     & processed data & \textbf{36.9}& \textbf{69.88} & \textbf{38.4}&  \textbf{69.04} & \textbf{36.94}& \textbf{69.37} & \textbf{37.40} & \textbf{69.58}\\
\hline
\multirow{2}{*}{PT + FT} & raw data &24.6 & 60.47& 27.13 & 59.14& 24.45 & 59.4& 26.1 & 58.7\\
                     & processed data & \textbf{38.22}& \textbf{72.09} & \textbf{39.98}&  \textbf{69.56} & \textbf{36.52}& \textbf{72.62} & \textbf{36.38} & \textbf{70.1}\\
\hline
\end{tabular}
\caption{\label{main_results}
Image-text retrieval results for the BLIP model were pre-trained/Fine-tuned (PT means only pre-training and PT + FT means pre-training plus fine-tuning) with two different schemes. One with following existing VL models and the other with following our pipeline. 
}
\end{table*}

\section{Brain Image-Caption for VL Pre-Training}

In this section, we explain the details of our pipeline for processing the data collected from PubMed and pre-training the vision-language (VL) model.

\subsection{Data collection and processing}
\paragraph{Image-caption pairs.} Out of 9,371 journal papers from 1,021 various journal titles like "Brain tumor research and treatment" and "BMC Medical Imaging" dated from 1937-2018, We collected image-caption pairs from PubMed and filtered them by only scraping "case reports" with both types of image and text data. Pairs are additionally filtered by the keyword "brain" that appears in the corresponding caption assuming that both the image and caption are brain-related data. We collected 22,795 data initially and pre-trained the BLIP model on image-caption pairs as our baseline. Other data from the medical domain can be collected following our procedure to build another domain-specific VL model. 

\begin{figure}[!ht]
\begin{center}
\includegraphics[scale=0.4]{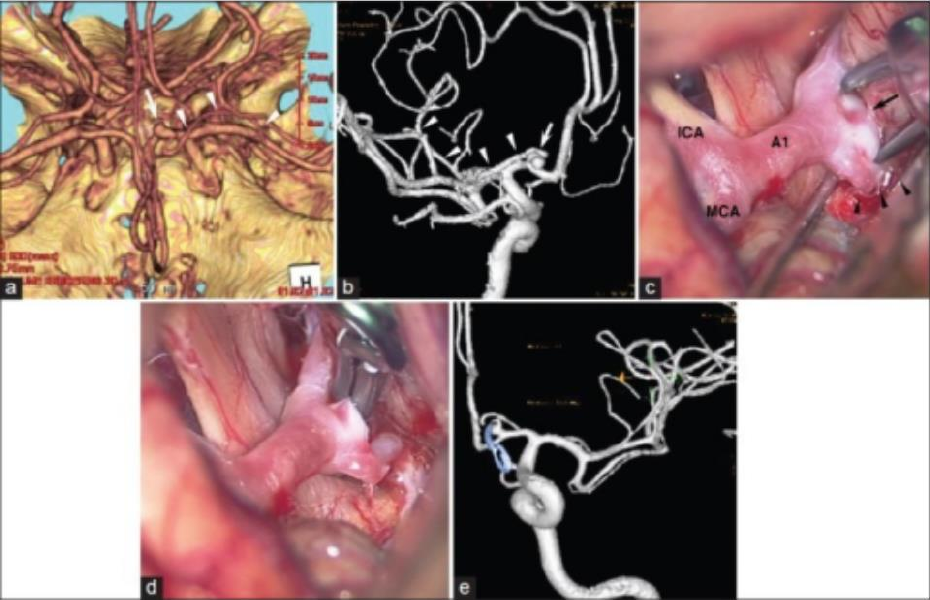} 
\caption{An example data from the dataset where each subfigure comes from a different source and Providing the model with the entire figure and its accompanying caption could potentially result in misdirection, causing it to emphasize image areas that are less relevant to both the caption and subcaptions.}
\label{image_data}
\end{center}
\end{figure}

\paragraph{Fine-grained alignment.} Due to the presence of subfigures and subcaptions in a significant portion of collected images ( ~43 percent of data ), there is a unique challenge in their usage for the pre-training process and following the pre-training schema of existing VL models (taking the entire image and entire caption as image-text pairs) could lead to a low-performance model. This is because the model may struggle to match subcaptions and subfigures, which is not the issue in general domain VL datasets. Additionally, in numerous images within the dataset, subfigures exhibit significant distribution variations, possibly stemming from diverse sources such as MRI scans, surgical cameras, or simulation visualizations (as exemplified in Fig. \ref{image_data}). Presenting the complete figure and caption to the model could potentially lead to misguidance, as it may focus on image portions less pertinent to the caption and subcaptions. Moreover, the complexity of medical terms in clinical notes and abnormal regions in images poses a significant challenge to the model's understanding. To overcome these challenges, we developed a pipeline shown in Figure 1 where we used an object detection model \cite{tsutsui2017data} to identify subfigures in the images while simultaneously leveraging an NLP tool \cite{wu2021melinda} to parse captions into subcaptions. With these two steps, we were able to convert the collected image-caption pairs into 39,535 subfigure/subcaption pairs. However, matching subfigures and subcaptions was another challenge in training the VL model. We utilized an OCR tool\footnote{google OCR: \url{https://cloud.google.com/vision/docs/ocr}} to detect the small characters in the subfigures commonly referred to as "subfigures labels" that are most likely the letters we can use to match with subcaptions. If the detected text does not match with any of the "subcaption labels" or does match with two or more of them, we pair the subfigure with the entire caption. Then aligned subfigures/subcaptions pairs are used for pre-training the model.

\section{Pre-Training and Evaluation} 

We trained the BLIP model on the processed data with a learning rate of 1e-5 for 60 epochs, using Adam as the optimization algorithm and Masked Language Modeling (MLM) \cite{devlin2018bert}, Image-text matching (ITM) \cite{chen2020uniter}, and Image-text contrastive (ITC) \cite{radford2021learning} as the pre-training objectives. We evaluated the pre-trained model on qualitative and quantitative intrinsic tasks including image-text retrieval. Image-text retrieval evaluates the model's ability to retrieve the corresponding text/image from a pool of data given an image/text. We also conducted qualitative analysis by visualizing the attention map of some important medical terms and their relationship with abnormal regions. 

\section{Results}

\subsection{Quantitaive evaluation}

\begin{figure}[!ht]
\begin{center}
\includegraphics[scale=0.25]{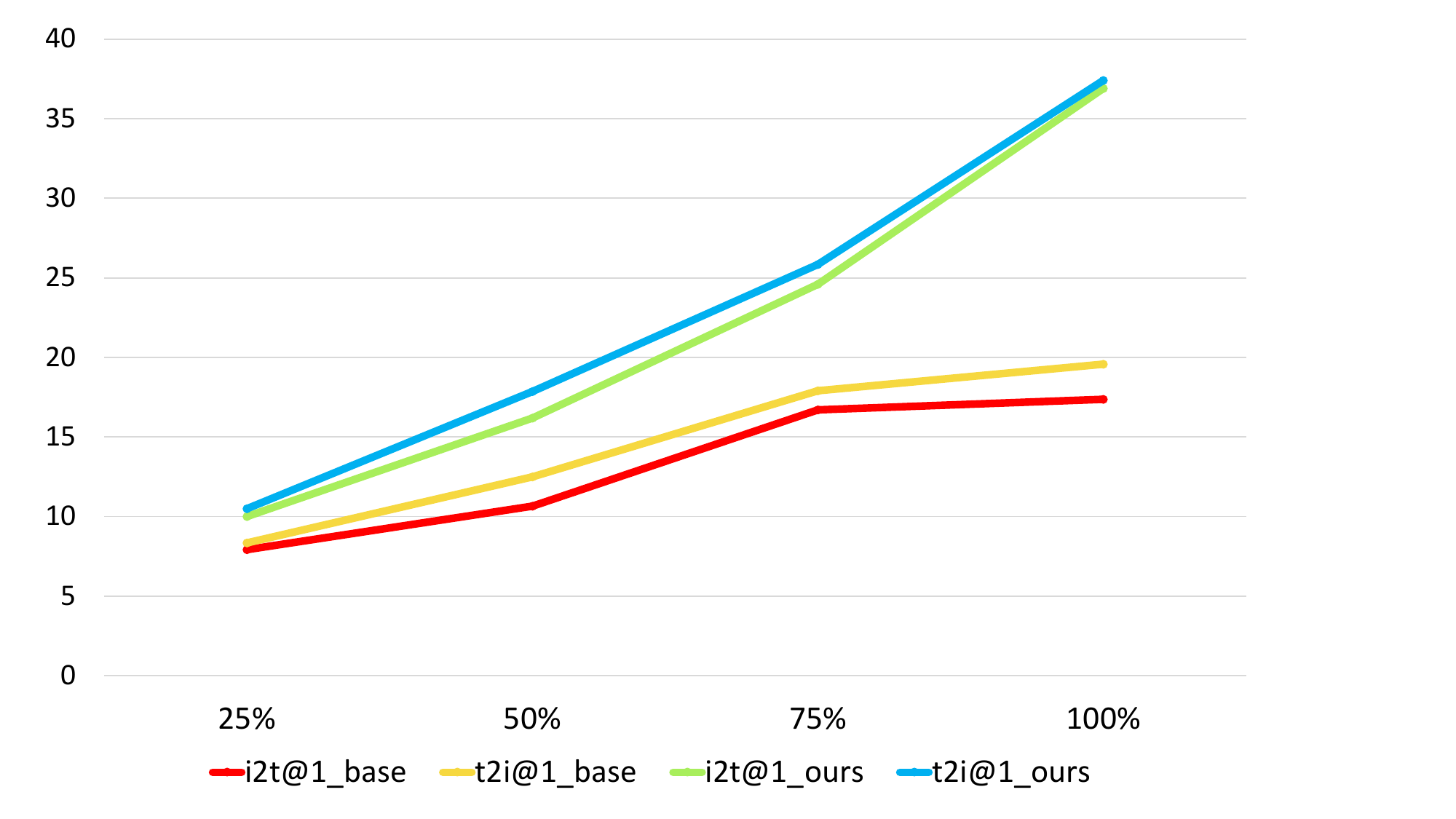} 

\caption{The plot visually demonstrates the dynamics of image-text retrieval performance across varying proportions of pre-training data. Our model results reveal a notable increase in learning capacity as more data becomes available, while the baseline exhibits characteristics that suggest a form of saturation.}
\label{quant}
\end{center}
\end{figure}

Our pre-trained model for image-text matching is evaluated to assess its ability to match an image to its corresponding caption. To perform this evaluation, we evaluate the model performance for image-text retrieval tasks as an intrinsic evaluation. In this task, we test how well our model can retrieve the corresponding caption given an image and a pool of all captions, and vice versa. We evaluate our model's performance in two settings: '@1', where we determine the percentage of val/test dataset where the corresponding image or caption is the top 1 retrieved data; and '@10', where we determine the percentage of data where the corresponding image or caption is among the top 10 retrieved data. 
Our results which are summarized in Table \ref{main_results}validate our pre-trained model's capability for VL modeling. 
Additionally, we conducted pre-training experiments with varying proportions of the available data, ranging from 25 to 100 percent, and assessed the impact on our image-text retrieval performance. As depicted in Figure \ref{quant}, our model exhibited a steeper performance curve, indicating its increased learning capacity as more data became available. In contrast, the baseline performance appeared to saturate and reach a plateau. This observation underscores the effectiveness of our pre-training pipeline in enhancing a model's learning capabilities. 

\subsection{Qualitative analysis}

\begin{figure}[!ht]
\begin{center}
\begin{overpic}
[width=0.45\textwidth]{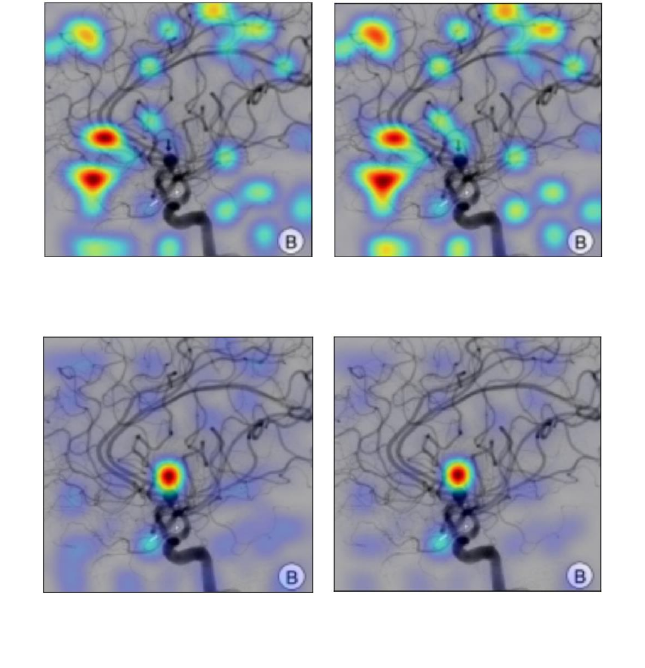} 
    \put(0,20){\rotatebox{90}{\small{{our model}}}}
    \put(0,70){\rotatebox{90}{\small{{baseline}}}}
    \put(20,5){\small{{cerebral}}}
    \put(65,5){\small{{artery}}}
    \put(20,55){\small{{cerebral}}}
    \put(65,55){\small{{artery}}}
\end{overpic}
\caption{The visualization of the attention of the "cerebral artery" on the corresponding image. It is evident that ours highlighted the abnormal region more specifically.}
\label{fig.2}
\end{center}
\end{figure}

\begin{figure}[!ht]
\begin{center}
\begin{overpic}
[width=0.45\textwidth]{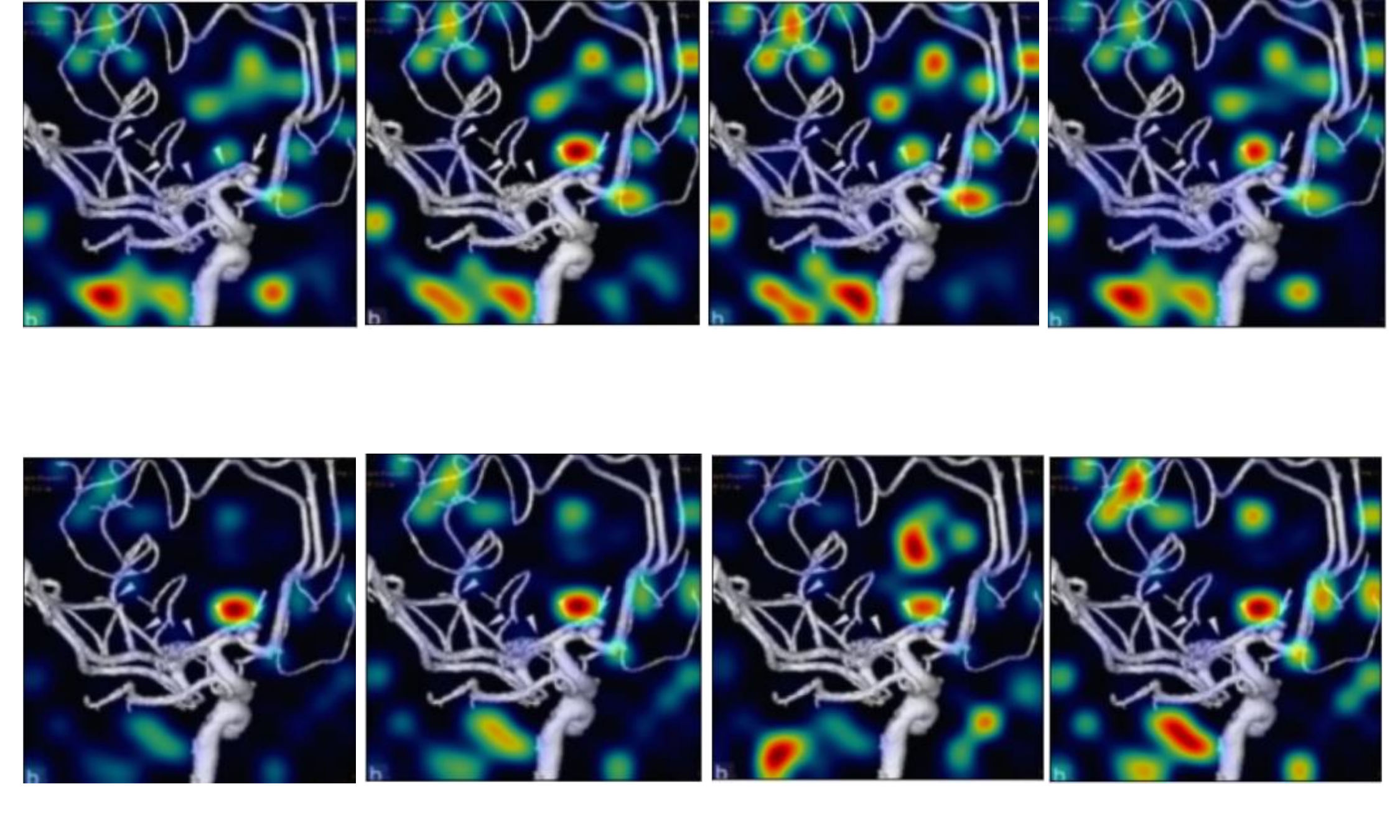}
    \put(-3,6){\rotatebox{90}{\small{{our model}}}}
    \put(-3,40){\rotatebox{90}{\small{{baseline}}}}
    \put(11,-1){\small{{an}}}
    \put(34,-1){\small{{\#eur}}}
    \put(59,-1){\small{{\#ys}}}
    \put(84,-1){\small{{\#m}}}
    \put(11,30){\small{{an}}}
    \put(34,30){\small{{\#eur}}}
    \put(59,30){\small{{\#ys}}}
    \put(84,30){\small{{\#m}}}
\end{overpic}
\caption{The visualization of the attention of the "aneurysm" on the corresponding image. It is evident that ours highlighted the abnormal region more specifically.}
\label{fig.3}
\end{center}
\end{figure}

As part of our qualitative analysis, we employ attention maps to visualize the model's focus on some important medical terms, such as 'aneurysm'—representing abnormal blood vessel outpouching—within the associated images and 'cerebral artery' a condition where there is a deviation from the normal, healthy state of arteries. In Fig. \ref{fig.2} and \ref{fig.3}, we present the model attention map for both the baseline and our pre-trained model concerning the terms 'cerebral artery' and 'aneurysm' respectively where 'aneurysm' is divided into subtokens. The upper heatmaps correspond to the baseline, while the lower ones depict the outputs of our pre-trained model. The attention maps demonstrate that our model emphasizes the relevant areas.

\section{Related Work}
\paragraph{Vision-Language Learning.} Large-scale VL pretraining has demonstrates impressive performance for various VL tasks~\cite{lu2019vilbert,tan-bansal-2019-lxmert,su2019vl,li2019visualbert}. Early works in this direction use off-the-shelf object detectors to extract objects~\cite{anderson2018bottom} and feed the region features into a modality fusion module~\cite{su2019vl,tan-bansal-2019-lxmert,chen2020uniter,li2020oscar,zhang2021vinvl}. End-to-end training methods have been proposed to be efficient and effective alternatives to object detector-based models~\cite{kim2021vilt,li2021align,dou2021empirical} because of their ability to unleash the power of vision encoders. While state-of-the-art VL models have achieved impressive performance, many of them are mainly focused on general-domain VL tasks such as visual question answering and image captioning, and it is unclear whether the pre-trained representations can be helpful for medical tasks.

\paragraph{Medical-Domain VL Learning.} Multi-modal learning in the medical domain is of great significance due to the presence of medical images, text notes, and electronic health records. While there are a few medical-domain datasets and models being proposed~\cite{eslami2023pubmedclip,liu2021contrastive}, many of them are focused on CLIP-style models~\cite{radford2021learning} trained with contrastive objectives, neglecting to investigate more effective VL training methods in a large-scale setting. In addition to this line of work, \citet{zhang2022mmformer} introduce the mmFormer, a novel Transformer-based approach, for accurate brain tumor segmentation from incomplete MRI modalities, achieving significant improvements in segmentation performance compared to state-of-the-art methods on the BraTS 2018 dataset. \citet{lin2023pmc} present PMC-OA, a large-scale biomedical dataset with 1.6M image-caption pairs from PubMedCentral's OpenAccess subset. Using this dataset, the PMC-CLIP model achieves state-of-the-art results in image-text retrieval and classification tasks, addressing data scarcity issues in the biomedical domain. In this work, we adapt the advanced techniques in the general domain VL learning to the medical domain and demonstrate its effectiveness.

\section{Conclusion}

Our study is centered on the creation of a pre-training dataset for domain-specific medical vision language models, utilizing a VL dataset sourced from PubMed and a pipeline for VL modeling. Our approach has been evaluated through intrinsic task assessments, demonstrating its effectiveness in constructing data-efficient VL models for the medical field. Our model has surpassed the performance of standard VL training methods, further validating the efficacy of our pipeline. Our proposed pipeline helps build powerful multimodal AI models which could be used for various tasks like diagnosis, drug discovery, and information extraction.

\section{Aknowledgment}

This research has received support from Optum and was partially funded by the National Institutes of Health (NIH) under grant number R01HL152270. We extend our gratitude to Joel Stremmel and his team at OptumLabs, as well as to the reviewers whose valuable feedback significantly contributed to the enhancement of our work.




%
\section{References}\label{reference}

\bibliographystyle{lrec-coling2024-natbib}
\bibliography{lrec-coling2024-example}

\end{document}